\documentclass{article}

\usepackage{microtype}
\usepackage{graphicx}
\usepackage{subfigure}
\usepackage{booktabs} 

\usepackage{hyperref}

\usepackage[accepted]{icml2025}

\usepackage{amsmath}
\usepackage{amssymb}
\usepackage{mathtools}
\usepackage{amsthm}

\usepackage[capitalize,noabbrev]{cleveref}

\theoremstyle{plain}

\theoremstyle{definition}

\theoremstyle{remark}

\usepackage[textsize=tiny]{todonotes}

\usepackage{bm}

\icmltitlerunning{Causality-Aware Contrastive Learning for Robust Multivariate Time-Series Anomaly Detection}

\begin{document}

\twocolumn[
\icmltitle{Causality-Aware Contrastive Learning for Robust\\ Multivariate Time-Series Anomaly Detection}



\icmlsetsymbol{equal}{*}

\begin{icmlauthorlist}
\icmlauthor{HyunGi Kim}{ece}
\icmlauthor{Jisoo Mok}{ece}
\icmlauthor{Dongjun Lee}{ai}
\icmlauthor{Jaihyun Lew}{ai}
\icmlauthor{Sungjae Kim}{hyundai}
\icmlauthor{Sungroh Yoon}{ece,ai,aiis}
\end{icmlauthorlist}

\icmlaffiliation{ece}{Department of Electrical and Computer Engineering, Seoul National University, Seoul, Republic of Korea}
\icmlaffiliation{ai}{Interdisciplinary Program in Artificial Intelligence, Seoul National University, Seoul, Republic of Korea}
\icmlaffiliation{aiis}{AIIS, ASRI, and INMC, Seoul National University, Seoul, Republic of Korea}
\icmlaffiliation{hyundai}{Hyundai Motor Company, Gyeonggi-do, Republic of Korea}

\icmlcorrespondingauthor{Sungroh Yoon}{sryoon@snu.ac.kr}

\icmlkeywords{Time-Series, Anomaly Detection, Causality, Contrastive Learning, Multivariate}

\vskip 0.3in
]



\printAffiliationsAndNotice{}  

\begin{abstract}
Utilizing the complex inter-variable causal relationships within multivariate time-series provides a promising avenue toward more robust and reliable multivariate time-series anomaly detection (MTSAD) but remains an underexplored area of research.
This paper proposes \textbf{C}ausality-\textbf{A}ware contrastive learning for \textbf{RO}bust multivariate \textbf{T}ime-\textbf{S}eries (CAROTS), a novel MTSAD pipeline that incorporates the notion of causality into contrastive learning.
CAROTS employs two data augmentors to obtain causality-preserving and -disturbing samples that serve as a wide range of normal variations and synthetic anomalies, respectively.
With causality-preserving and -disturbing samples as positives and negatives, CAROTS performs contrastive learning to train an encoder whose latent space separates normal and abnormal samples based on causality.
Moreover, CAROTS introduces a similarity-filtered one-class contrastive loss that encourages the contrastive learning process to gradually incorporate more semantically diverse samples with common causal relationships. 
Extensive experiments on five real-world and two synthetic datasets validate that the integration of causal relationships endows CAROTS with improved MTSAD capabilities. 
The code is available at \href{https://github.com/kimanki/CAROTS}{https://github.com/kimanki/CAROTS}.
\end{abstract}

\section{Introduction} 
Multivariate time-series anomaly detection (MTSAD) aims to identify anomalies in time-series collected across multiple interdependent variables~\cite{li2023deep, choi2021deep}. 
Applications of MTSAD are far-reaching, spanning from detecting network intrusion in cyber-physical systems~\cite{cps1} to monitoring patient vitals in healthcare systems~\cite{healthcare1, healthcare2}. 
Since undetected anomalies can cause serious harm in mission-critical systems, MTSAD plays a crucial role in ensuring their operational efficiency and security.
Furthermore, as such systems continue to grow in size and complexity, with an increasing number of variables, understanding the intricate inter-variable relationships rises as a critical challenge in MTSAD~\cite{kang2024transformer}.

Causality offers a structured approach to capture the underlying relationships within multivariate time-series~\cite{runge2019detecting}.
In multivariate time-series, the observed variables are governed by causal relationships that dictate how changes in one variable affect others~\cite{granger1969investigating, cheng2024causaltime}. 
For instance, the increase in temperature directly causes the duration for which the air conditioner stays on to be longer.
Once an anomaly arises, provoking an unexpected behavior from one or more variables, these causal relationships break down (\textit{e.g.}, a broken air conditioner will remain turned off regardless of the temperature). 
Except in the case of such an abnormal event, the inter-variable causal relationships generally remain consistent, and thus, a violation of causal relationships can be used as a key indicator of anomalies~\cite{liu2025gcad, febrinanto2023entropy}. 

Understanding the complex inter-variable dynamics from the perspective of causality enables a more reliable and robust MTSAD by allowing discernment of causality-preserving variations of normal instances (\textit{e.g.,} the air-conditioner staying on for longer than usual due to heatwaves) from true anomalies with broken causality.
However, current unsupervised MTSAD methods, such as reconstruction- or contrastive learning-based approaches, focus instead on superficial differences in data values or distributions and overlook inter-variable causal relationships~\cite{SARAD, AnomalyTransformer, kim2023contrastive}. 
Failing to account for inter-variable causal relationships can lead to misidentifying normal variations as anomalies, resulting in false alarms and reduced detection accuracy. 

In this work, we propose a novel causality-aware MTSAD framework dubbed CAROTS, which is a shorthand for \textbf{C}ausality-\textbf{A}ware contrastive learning for \textbf{RO}bust M\textbf{TS}AD.
CAROTS first extracts the causal relationships among normal multivariate time-series by leveraging a forecasting-based causal discovery model. 
Afterward, it augments the training data with both causality-preserving and -disturbing samples. 
The former represents a diversified set of normal, causality-preserving operations, while the latter simulates anomalies that violate causal relationships. 
To explicitly separate normal operations from anomalies based on the causal structure, CAROTS utilizes contrastive learning, where causality-preserving and -disturbing samples serve as positive and negative examples, respectively.
In the embedding space of this contrastively-trained encoder, the causality-preserving and -disturbing samples form their own clusters that are clearly distinguishable from each other.

In addition, we introduce a similarity-filtered one-class contrastive loss, which guides the contrastive learning process to gradually incorporate more semantically diverse samples that share common normal causal relationships.
With each sample in a mini-batch as an anchor, the similarity filter excludes low-similarity positive samples, which fall below a pre-set threshold, from the loss computation.
Thus, the similarity filter only treats samples with highly similar patterns as positives in early iterations.
As training progresses, positive samples with more diverse patterns are incorporated, allowing the trained encoder to include a broader range of normal variations.
Finally, CAROTS performs anomaly detection by ensembling two causality-aware anomaly scores: the distance between a test time-series and the centroid of normal, causality-preserving embeddings obtained with the contrastively-trained encoder and the forecasting error of the forecasting-based causal discovery model.

We validate the effectiveness of CAROTS across five real-world MTSAD datasets, where it consistently outperforms existing MTSAD methods. 
On two synthetic datasets with explicit linear and non-linear causal relationships, CAROTS again significantly outperforms methods that overlook causal relationships. 
In particular, CAROTS successfully detects even the most difficult synthetic anomalies, which other baselines all fail to identify; the remarkable improvement in AUROC up to 50\% verifies that the integration of causal knowledge endows CAROTS with the ability to reliably detect anomalies.
We demonstrate that CAROTS is robust to changes in hyperparameters and encoder architecture to showcase its practical applicability. 
Lastly, the ablation study on each technical component of CAROTS verifies its individual contributions.

We summarize our contributions as follows:
\begin{itemize}
\item{We propose CAROTS, a novel MTSAD pipeline that leverages causality-preserving and -disturbing augmentors to induce causal-awareness in contrastive learning. The trained encoder successfully separates normal and abnormal samples based on causal relationships.}
\item{We propose a similarity-filtered one-class contrastive loss, which enforces the latent space to respect diverse semantic patterns by progressively incorporating more dissimilar samples throughout training.}
\item{An extensive empirical verification of CAROTS on five real-world and two synthetic datasets highlights its outstanding ability to detect anomalies, especially those that previous baselines thoroughly struggle with.}
\end{itemize}

\begin{figure*}[t]
    \centering
    \includegraphics[width=\linewidth]{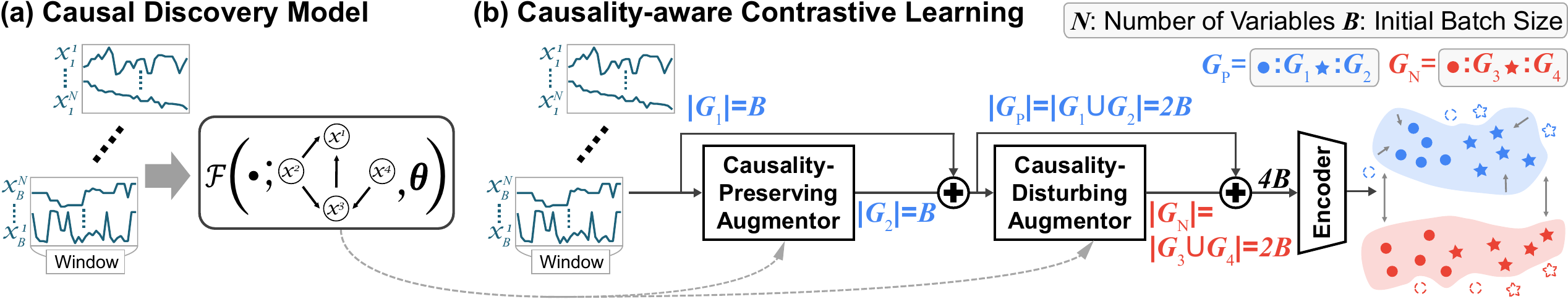}
    \vspace{-1.5em}
    \caption{An overall pipeline of CAROTS. \textbf{(a)} First, a causal discoverer is trained to learn the normal causal relationships in the training data. \textbf{(b)} Based on the causal discoverer, the causality-preserving and -disturbing augmentors construct a mini-batch of four groups of samples: original ($G_1$), causality-preserving augmentations ($G_2$), causality-disturbing augmentations of $G_1$ ($G_3$), and causality-disturbing augmentations of $G_2$ ($G_4$). Contrastive learning with the similarity-filtered one-class contrastive loss clusters positive groups ($G_P \coloneqq G_1 \cup G_2$) while separating them from ($G_N \coloneqq G_3 \cup G_4$) in the embedding space. $N$, $B$, and $\oplus$ denote the number of variables, initial batch size, and concatenation operation, respectively.}
    \label{fig:overview}
\end{figure*}

\section{Related Works}
\subsection{Multivariate Time-series Anomaly Detection}
\textbf{Reconstruction-based methods. }
Reconstruction-based MTSAD methods train a reconstruction model exclusively on normal data~\cite{wang2025survey, SARAD}.
Anomalies are detected based on the reconstruction error of this model, under the assumption that they will yield higher reconstruction errors.
This line of research mostly focuses on utilizing advanced architectures. 
AnomalyTransformer~\cite{AnomalyTransformer} introduces anomaly attention which models prior and series associations to distinguish normal and anomalous points. 
TimesNet~\cite{TimesNet} exploits TimesBlock that learns multi-scale temporal dependencies based on Fourier transform. 
Aside from studying architectural variants, USAD~\cite{USAD} uses adversarial training to build a more reliable reconstruction model.
However, the sole reliance on normal training samples limits their ability to fully consider the variability of normal operations or integrate insights related to potential anomalies. 

\textbf{Contrastive learning-based methods. }
Contrastive learning is a self-supervised learning approach that aims to learn latent representations of data by contrasting positive and negative sample pairs~\cite{SimCLR, moco, lee2024soft}. 
Its usefulness in anomaly detection has been shown across various domains~\cite{csi, ssd}. 
In anomaly detection, contrastive learning is used to train an encoder that can distinguish normal and abnormal patterns in a self-supervised manner.
The negative samples in contrastive MTSAD allow the integration of simulated anomalies during the training phase, unlike in reconstruction-based methods~\cite{darban2025carla, kim2023contrastive}.
The efforts to advance contrastive MTSAD can be categorized into: data augmentation techniques~\cite{ky2024cats, choi2024self}, integration of one-class learning~\cite{kim2023contrastive, xu2024calibrated}, and self-supervised classification~\cite{yang2023dcdetector, ngu2023cl}.  

Our work sets itself apart from previous contrastive MTSAD by integrating the concept of causality into the data augmentation process. 
In contrastive MTSAD, data augmentation functions that reflect the traits of time-series are preferable for creating positive and negative examples that simulate normals and abnormal samples.
In this vein, CL-TAD~\cite{ngu2023cl} leverages encoder-decoder-based reconstruction with randomly masked inputs, while CTAD~\cite{kim2023contrastive} offers guidelines for using general time-series augmentations for contrastive MTSAD. 
CARLA~\cite{darban2025carla} introduces various types of synthetic anomalies, such as point, contextual, and collective anomalies, to improve anomaly detection. 
In contrast, our approach employs causality-preserving and causality-disturbing augmentors, allowing the model to explicitly exploit causal relationships, an essential characteristic of multivariate time-series.

\subsection{Causal Discovery in Multivariate Time-series}
Causal discovery uncovers causal relationships among interdependent variables in multivariate time-series~\cite{assaad2022survey, cheng2024causaltime}. 
The Granger causality test~\cite{granger1969investigating} has been widely adopted to infer causal relationships by testing whether past values of one variable improve the prediction of another. 
Recently, deep learning approaches have been used for more robust causal discovery~\cite{tank2021neural, yuxiao2023cuts, cheng2024cuts+}. 
NGC~\cite{tank2021neural} infers non-linear causal relationships in multivariate time-series by combining structured neural networks with sparsity-inducing regularization. 
CUTS~\cite{yuxiao2023cuts} jointly imputes missing data and constructs causal graphs, and CUTS+~\cite{cheng2024cuts+} extends this by improving scalability for high-dimensional data with a Coarse-to-fine Discovery technique and a Message-Passing-based Graph Neural Network.
Causalformer~\cite{kong2024causalformer} integrates causal discovery into Transformer, learning sparse adjacency matrices to capture temporal and causal relationships. 
Contrary to the exhaustive efforts toward causal discovery in time-series, leveraging causal relationships to distinguish anomalies from normal operations in MTSAD remains underexplored~\cite{qiu2012granger, xing2023gcformer, liu2025gcad}.

\section{Preliminaries}
\label{sec:preliminary}
\subsection{Causal Discovery Model}
\label{subsec:causal_discovery}
A multivariate time-series is defined as a collection of time-series obtained from $N$ variables $\{x^i\}_{i=1}^N$, where $N > 1$. 
According to the Granger causality~\cite{granger1969investigating}, $x^i$ causes $x^j$ if the past values of $x^i$ affect the future values of $x^j$. 
In this case, $x^i$ is the cause of $x^j$ while $x^j$ is the effect of $x^i$. 
We denote a set of causes and effects of $x^i$ as $\mathrm{Pa}(x^i)$ and $\mathrm{Ch}(x^i)$, respectively. 
The objective of a causal discovery model is to search for $\mathrm{Pa}(x^i)$ and $\mathrm{Ch}(x^i)$ for every $x^i$ and to learn the function $f^i$ that maps the causal relationships between $x^i$ and $\mathrm{Pa}(x^i)$.
Given $\mathrm{Pa}(x^i)$ and $f^i$, the value of $x^i$ at time $t$, denoted as $x^i_t$, can be represented as $x^i_t = f^i(\bm{x}^i_{<t}, \mathrm{Pa}(x^i)_{<t})$ where $\bm{x}^i_{<t}$ denotes the values of $x^i$ preceding $t$. 

The forecasting-based causal discovery model $\mathcal{F}_{\bm{\theta}, \bm{A}}(\cdot)$ represents the causal relationships with the causality matrix $\bm{A} \in \{0, 1\}^{N \times N}$ and causal relationship functions $\{f^i(\cdot)\}_{i=1}^{N}$. 
The causality matrix $\bm{A}$ represents $\mathrm{Pa}(x^i)$ and  $\mathrm{Ch}(x^i)$, where $\bm{A}_{i, j} = 1$ holds if and only if $x^i \in \mathrm{Pa}(x^j)$.
To train $\mathcal{F}_{\bm{\theta}, \bm{A}}(\cdot)$, a single, contiguous multivariate time-series $\{\bm{x}_t\}_{t=1}^T$ is segmented with a sliding window with window size $w$ such that $\bm{X}_{t} = \{ \bm{x}_{t-w+1}, \dots, \bm{x}_t\} \in \mathbb{R}^{w \times N}$, where $\bm{x}_t \in \mathbb{R}^N$ denotes the values of $N$ variables at time $t$. 
$\mathcal{F}_{\bm{\theta}, \bm{A}}(\cdot)$ is trained to minimize the forecasting Mean Squared Error (MSE) as follows:
\begin{equation}
\hat{\bm{\theta}}, \hat{\bm{A}} = \operatorname*{\mathrm{argmin}}\limits_{\bm{\theta}, \bm{A}} \operatorname*{\mathbb{E}}\limits_{\bm{X}_t} \left[ \mathrm{MSE} \left( \mathcal{F}_{\bm{\theta}, \bm{A}}\left(\bm{X}_{<t}\right), \bm{x}_t \right) + \lambda ||\bm{A}||_1 \right],
\end{equation}
where $\bm{X}_{<t}$ denotes values in $\bm{X}_{t}$ preceding $t$. 
$\mathcal{F}_{\bm{\theta}, \bm{A}}(\cdot)$ approximates $\{f^i(\cdot)\}_{i=1}^{N}$ that predicts $\bm{x}_{t}$ based on the preceding values $\bm{X}_{<t}$ and $\bm{A}$.
Our work utilizes CUTS+~\cite{cheng2024cuts+}, a representative forecasting-based causal discovery model, to extract meaningful causal structures. 

\subsection{MTSAD through the Lens of Causality}
Following the general assumption in time-series analysis~\cite{absar2021discovering, absar2023neural}, we assume the causal relationships in normal multivariate time-series remain consistent over time (\textit{i.e.,} $\bm{A}$ and $\{f^i (\cdot) \}_{i=1}^{N}$ are time-invariant). From the causal perspective, multivariate time-series anomalies can then be interpreted as values that deviate from the normal causal relationships:
\begin{equation}
\exists x_t^i \, : \, | x^i_t - f^i(x^i_{<t}, \mathrm{Pa}(x^i)_{<t})| > \delta.
\label{eq:loss_cd}
\end{equation}
In this study, we harness this causal interpretation of anomalies by integrating the notion of causality into contrastive MTSAD. 
By exploiting synthetic samples that either preserve normal, consistent causal relationships or violate them as in Eq.~\ref{eq:loss_cd}, we enable an explicitly causal-aware discrimination of anomalies in MTSAD.

\begin{figure*}[t]
    \centering
    \includegraphics[width=\linewidth]{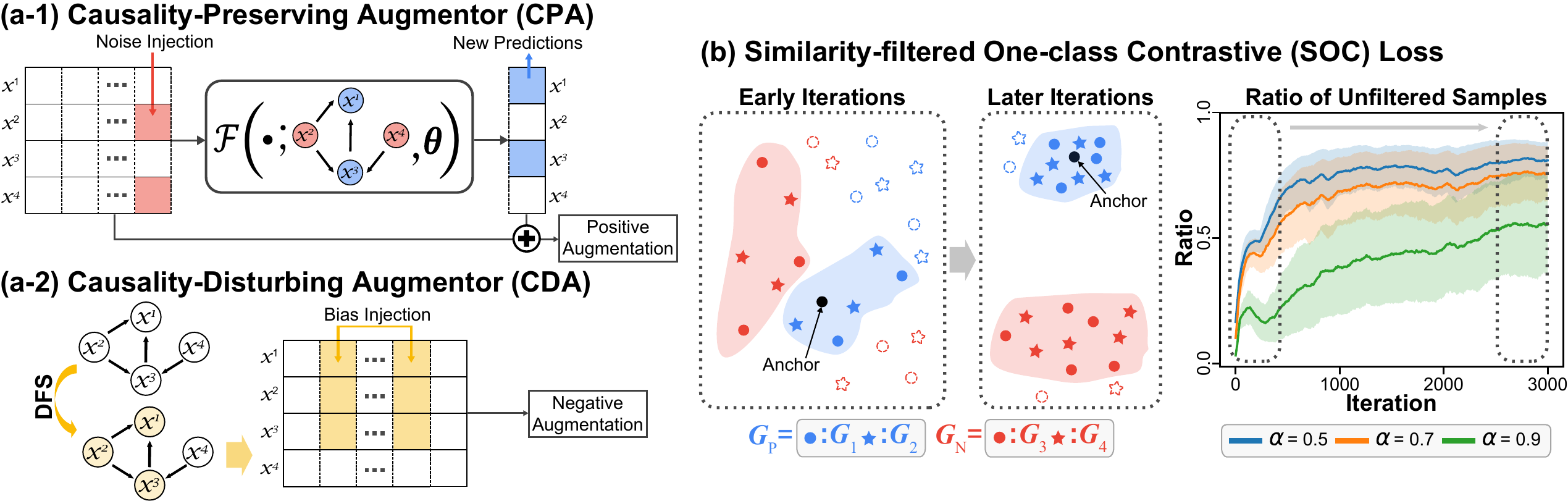}
    \vspace{-1.5em}
    \caption{\textbf{(a-1)} CPA adds Gaussian noise to randomly selected variables (\textcolor{red}{Red}) and uses the causality-integrated forecaster to predict and replace the values of affected variables (\textcolor{blue}{Blue}). \textbf{(a-2)} CDA selects a random variable (\textcolor{yellow}{Yellow}) and extracts a directed subgraph of a causality matrix through DFS. Random perturbations that are unrelated to causal relationships are injected into the selected variables.\textbf{(b)} [Left] In early iterations, SOC loss filters out low-similarity positive samples (Dashed) based on a threshold $\alpha$, ensuring that the loss is applied only to high-similarity positive samples and their causality-disturbed counterparts (Solid-colored). As training progresses, more diverse samples engage in the SOC loss computation. [Right] shows that a ratio of unfiltered samples steadily increases during training.}
    \label{fig:components}
\end{figure*}

\section{Methodology}
We now introduce CAROTS, \textbf{C}ausality-\textbf{A}ware contrastive learning for \textbf{RO}bust M\textbf{TS}AD. 
Section~\ref{sec:augmentation} discusses how the causal relationships, extracted by a causal discoverer, are employed for causality-aware data augmentation. 
Section~\ref{sec:loss} elaborates how the causality-informed discriminative embedding space is obtained through contrastive learning with the proposed similarity-filtered one-class contrastive loss.
Lastly, Section~\ref{sec:score} describes how the anomalies are scored based on the overall framework.
\figurename~\ref{fig:overview} illustrates the overall pipeline of CAROTS.

\subsection{Causality-aware Data Augmentation}
\label{sec:augmentation}
The process in Section~\ref{sec:preliminary} yields a causality matrix $\bm{A}$, which encodes the causal interactions among variables, and a causality-integrated forecaster $\mathcal{F}_{\bm{\theta}, \bm{A}}(\cdot)$ that predicts future values given these causal relationships.
Because unsupervised MTSAD assumes that training data are void of anomalies, the learned $\bm{A}$ and $\mathcal{F}_{\bm{\theta}, \bm{A}}(\cdot)$ represent normal causal relationships.
With $\bm{A}$ and $\mathcal{F}_{\bm{\theta}, \bm{A}}(\cdot)$, CAROTS generates causality-preserving and -disturbing samples with respective augmentors, visualized in~
\figurename~\ref{fig:components}.

\noindent\textbf{Causality-Preserving Augmentor (CPA)} generates positive samples with intact causal relationships. 
Causality-preserving samples from CPA increase the diversity of normal data that maintain causal relationships and simulate wide-ranging normal data not included in the training set.
For a given input $\bm{X}_t$, CPA randomly selects a set of $M < N$ causing variables $\bm{C}$ such that $\bm{C} \subset \{x^i\}_{i=1}^{N}$, $|\bm{C}| = M$.
Then, a set of variables directly affected by $\bm{C}$, defined as $\bm{E} \coloneqq \{ \mathrm{Ch}(x^j) \, | \, x^j \in \bm{C} \}$, is identified based on $\bm{A}$.
CPA adds small Gaussian noise $z \sim \mathcal{N}(0, \sigma^2)$ to trigger variations in the causing variables.
The modified causing variables are fed into $\mathcal{F}_{\bm{\theta}, \bm{A}}(\cdot)$, which predicts new values of $\bm{E}$ according to the learned causal relationship.
By replacing the original values of $\bm{E}$ with the predictions of $\mathcal{F}_{\bm{\theta}, \bm{A}}(\cdot)$, CPA obtains new causality-preserving samples that maintain the original causal relationships as the following:

\begin{equation}
\begin{aligned}
\left( \bm{X'}_{<t} \right)^i &= 
\begin{cases}
\left( \bm{X}_{<t} \right)^i + z \text{\,\,\, if \,} \bm{x}^i \in \bm{C} \\
\left( \bm{X}_{<t} \right)^i \text{\,\,\, otherwise}, 
\end{cases} \\
(\bm{x'}_t)^j &= 
\begin{cases}
    \left( \mathcal{F}_{\bm{\theta}, \bm{A}}(\bm{X'}_{<t}) \right)^j \text{\,\,\, if \,} \bm{x}^j \in \bm{E} \\
    \bm{x}_t^j \text{\,\,\, otherwise},
\end{cases} \\
\mathrm{CPA}(\bm{X}_t) &= \mathrm{Concat}\left( \{ \bm{X'}_{<t}, \bm{x'}_t \} \right).
\end{aligned}
\end{equation}

\noindent\textbf{Causality-Disturbing Augmentor (CDA)} generates negative samples with broken causal relationships.
Causality-disturbing samples from CDA act as proxies for anomalies caused by disrupted causal relationships.
Multivariate time-series can be represented in a directed graph, whose nodes are all variables and edges are formed according to $\bm{A}$.
Like in CPA, CDA first selects a random set of causing variables $\bm{C}$.
Beginning from $\bm{C}$, CDA performs a depth-first search (DFS) to generate multiple subgraphs as shown in \figurename~\ref{fig:components}.
Each iteration of DFS terminates with a cut-off probability of $p$, allowing subgraphs of varying sizes to be extracted.
CDA perturbs the variables in the extracted subgraph by injecting random biases so that the original causal relationships no longer hold.

\subsection{Causality-aware Contrastive Learning with Similarity-filtered One-class Contrastive Loss}
\label{sec:loss}
With samples from CDA and CPA, CAROTS performs contrastive learning to train an encoder $\mathcal{E}_{\bm{\phi}}(\cdot)$ that explicitly separates normal, causality-preserving samples from abnormal, causality-disturbing samples.
In the latent space of the resulting encoder, trained with causality-aware contrastive learning, the embeddings of causality-preserving and -disturbing samples appear far apart from each other.

Given a mini-batch of $B$ samples ($G_1$), the CPA first generates $B$ additional causality-preserving samples ($G_2$). 
Afterward, the CDA is applied simultaneously to $G_1$ and $G_2$ to obtain causality-disturbing samples from both original and causality-preserving augmented samples. 
As a result, CDA yields a total of 2$B$ causality-disturbing samples.
The four types of data -- $G_1$, $G_2$, $B$ causality-disturbing samples from original samples ($G_3$), and $B$ causality-disturbing samples from causality-preserving augmented samples ($G_4$) -- form an augmented mini-batch $\mathcal{B}$:  
\begin{equation}
\mathcal{B} = \mathrm{Concat}\left(\{ G_1, G_2, G_3, G_4 \}\right), \,\, |\mathcal{B}| = 4B.
\end{equation}

These four groups of data are used for positive and negative samples in contrastive learning as:
\begin{equation}
\begin{aligned}
G_P &\coloneqq G_1 \cup G_2 \\
G_N &\coloneqq G_3 \cup G_4,
\end{aligned}
\end{equation}
where $G_P$ and $G_N$ denote a set of positive and negative samples, respectively.
$G_P$ allows the encoder to effectively embed a wide variety of causality-preserving samples, including those not immediately available as a part of the original training data.
$G_N$ enables the encoder to distinguish anomalies caused by disturbances in causal relationships.
In particular, we note that $G_2$ and its causality-disturbing counterparts $G_4$ equip the encoder to generalize to normal variations and anomalies that arise from them.

In $G_P$, semantically diverse samples that share common causal relationships exist (\textit{e.g.} the causal relationship between the temperature and the air conditioner usage is preserved, but diverse patterns are generated depending on the air conditioner setting, such as the wind strength).
Therefore, $G_P$ can be grouped into multiple clusters based on the semantic similarity of individual samples. 
However, the na\"ive one-class contrastive loss forces all samples in the $G_P$ to collapse into a single cluster, which in turn prevents the encoder's embedding space from encoding the semantic diversity of samples belonging to $G_P$.

To address this pitfall, we introduce~\textbf{Similarity-filtered One-class Contrastive Loss} (SOC), whose overall scheme is illustrated in~\figurename~\ref{fig:components}.
At each update step, SOC filters out positive samples in $G_P$ whose similarity with an $i$-th anchor sample is lower than a pre-defined threshold $\alpha$. 
Then, SOC applies the one-class contrastive loss only to the remaining positive and corresponding negative samples, whose indices are denoted as $P_i$ and $N_i$, respectively.
The proposed SOC is formulated as follows: 
\begin{equation}
\begin{aligned}
\mathcal{L} = \frac{1}{2B} \sum_{i=1}^{2B} &\frac{1}{|P_i|} \sum_{j \in P_i} -\log \frac{\exp(\mathcal{S}_{i, j})}{\exp(\mathcal{S}_{i, j}) + \sum\limits_{k \in N_i} \exp(\mathcal{S}_{i, k})}, \\
P_i &= \{ j \in \{1, 2, \dots, 2B\} \mid \mathcal{S}_{i, j} \geq \alpha / \tau \}, \\
N_i &= \{ j + 2B \mid j \in P_i \},
\end{aligned}
\end{equation}
where $\mathcal{S}_{i, j}$ denotes cosine similarity between embeddings of $i$-th and $j$-th sample in $\mathcal{B}$ divided by temperature $\tau$.
SOC encourages each anchor sample to be aggregated with semantically similar samples first, allowing the encoder to respect different semantic clusters within $G_p$. 
As positive embeddings are aggregated during training, the proportion of filtered positive samples decreases, gradually incorporating more diverse samples sharing causal relationships.

\subsection{Anomaly Score}
\label{sec:score}
CAROTS conducts anomaly detection by combining two causality-aware anomaly scores: the distance between a test time-series and the centroid of normal, causality-preserving data embeddings obtained with the contrastively-trained encoder ($\mathcal{A}_{\mathrm{CL}}$) and the forecasting error of the forecasting-based causal discovery model ($\mathcal{A}_{\mathrm{CD}}$).

To compute $\mathcal{A}_{\mathrm{CL}}$, CAROTS relies on the latent space of the contrastively-trained encoder $\mathcal{E}_{\bm{\phi}}(\cdot)$ from Section~\ref{sec:loss}.
Given this latent space, CAROTS obtains the centroid $\bm{\mu}_P$ of embeddings of all positive samples, including original data.
Since the encoder is trained to cluster the embeddings of $G_P$ closely while pushing them away from those of causality-disturbing $G_N$, the distance between a test time-series and $\bm{\mu}_P$ is indicative of how much the test time-series deviates from causality-preserving positive samples. 
Consequently, larger $\mathcal{A}_{\mathrm{CL}}$ signifies
 that the corresponding test time-series is more likely to be anomalies.

CAROTS additionally incorporates $\mathcal{A}_{\mathrm{CD}}$, the forecasting error of forecasting-based causal discovery model $\mathcal{F}_{\bm{\theta}, \bm{A}}(\cdot)$ in Section~\ref{subsec:causal_discovery}, as an auxiliary anomaly score. 
$\mathcal{F}_{\bm{\theta}, \bm{A}}(\cdot)$, which is trained to predict future values based on learned normal causal relationships, struggles to forecast future values of abnormal samples that deviate from normal causal relationships.
Therefore, the forecasting error of $\mathcal{F}_{\bm{\theta}, \bm{A}}(\cdot)$ provides another crucial signal for causality-driven detection of anomalies following Eq.~\ref{eq:loss_cd}. 
Like $\mathcal{A}_{\mathrm{CL}}$, test-time time-series with large $\mathcal{A}_{\mathrm{CD}}$ are considered to be anomalies.

To ensemble the two scores, CAROTS normalizes each score to match the scales. 
Specifically, we calculate the mean and standard deviation of each score over the training data and apply z-normalization. 
The final anomaly score is obtained by summing the two z-scores as the following:
\begin{equation}
\begin{split}
\mathcal{A}(\bm{X}_t) &= \mathcal{A}_{\textrm{CL}}^{\textrm{norm}}(\bm{X}_t) + \mathcal{A}_{\textrm{CD}}^{\textrm{norm}}(\bm{X}_t), \\
\mathcal{A}_{\textrm{CL}}(\bm{X}_t) &= \mathcal{D}\left( \mathcal{E}_{\bm{\phi}}(\bm{X}_t), \bm{\mu}_P\right),\\
\mathcal{A}_{\textrm{CD}}(\bm{X}_t) &= \mathrm{MSE}\left(\mathcal{F}_{\bm{\theta}, \bm{A}}\left(\bm{X}_{<t}\right), \bm{x}_{t}\right),
\end{split}
\end{equation}
where $\mathcal{A}_{\textrm{CL}}^{\textrm{norm}}$ and $\mathcal{A}_{\textrm{CD}}^{\textrm{norm}}$ denote the z-normalized $\mathcal{A}_{\textrm{CL}}$ and $\mathcal{A}_{\textrm{CD}}$, respectively, and $\mathcal{D}$ denotes distance function.

\begin{table*}[t]
\footnotesize
\centering
\caption{Evaluation on widely-used real-world MTSAD datasets. AT and TN denote AnomalyTransformer and TimsNet, respectively. CAROTS$^\dag$ refers to the CAROTS without ensemble scoring, which only employs contrastive learning-based anomaly score.}
\label{tab:main}
\vspace{0.5em}
\begin{tabular}{ll||cccccccc||cc}
\toprule
    Dataset     & Metric & AT & TN & USAD & SimCLR & SSD & CSI & CTAD & CARLA & CAROTS & CAROTS$^\dag$ \\
          \midrule
                & AUROC & 0.501 & 0.808 & 0.812 & 0.528 & 0.486 & 0.540 & 0.820 & 0.807 & \underline{0.852} & \textbf{0.861} \\
    SWaT        & AUPRC & 0.477 & 0.713 & 0.720 & 0.136 & 0.264 & 0.146 & 0.702 & 0.691 & \textbf{0.764} & \underline{0.760} \\
                & F1    & 0.222 & 0.762 & 0.755 & 0.289 & 0.394 & 0.297 & 0.755 & 0.742 & \textbf{0.791} & \underline{0.789} \\
          \midrule
                & AUROC & 0.430 & 0.493 & 0.493 & 0.519 & 0.471 & 0.512 & \underline{0.599} & 0.533 & 0.502 & \textbf{0.622} \\
    WADI        & AUPRC & 0.045 & 0.055 & 0.054 & 0.106 & 0.246 & \underline{0.297} & \textbf{0.327} & 0.103 & 0.056 & 0.260 \\
                & F1    & 0.116 & 0.140 & 0.133 & 0.183 & 0.130 & 0.141 & \underline{0.374} & 0.175 & 0.143 & \textbf{0.391} \\
          \midrule
                & AUROC & 0.500 & 0.742 & 0.694 & \underline{0.766} & 0.689 & 0.695 & 0.708 & 0.445 & \textbf{0.783} & 0.729 \\
    PSM         & AUPRC & \textbf{0.642} & 0.547 & 0.538 & 0.593 & 0.514 & 0.511 & 0.527 & 0.257 & \underline{0.595} & 0.535 \\
                & F1    & 0.443 & \underline{0.589} & 0.506 & 0.554 & 0.498 & 0.498 & 0.515 & 0.444 & \textbf{0.603} & 0.534 \\
          \midrule
                & AUROC & 0.503 & 0.702 & 0.707 & 0.655 & 0.602 & 0.662 & \textbf{0.744} & 0.546 & 0.703 & \underline{0.726} \\
    SMD\_2-1    & AUPRC & \textbf{0.530} & 0.321 & \underline{0.349} & 0.227 & 0.184 & 0.231 & 0.229 & 0.156 & 0.184 & 0.193 \\
                & F1    & 0.103 & \underline{0.353} & \textbf{0.390} & 0.240 & 0.207 & 0.242 & 0.254 & 0.202 & 0.260 & 0.299 \\
          \midrule
                & AUROC & 0.584 & 0.767 & 0.727 & 0.642 & 0.640 & 0.664 & 0.659 & 0.483 & \underline{0.769} & \textbf{0.779} \\
    SMD\_3-7    & AUPRC & 0.225 & 0.340 & 0.300 & 0.101 & 0.044 & 0.251 & 0.120 & 0.171 & \textbf{0.430} & \underline{0.392} \\
                & F1    & 0.190 & 0.436 & 0.380 & 0.197 & 0.093 & 0.329 & 0.193 & 0.254 & \textbf{0.564} & \underline{0.542} \\
          \midrule
                & AUROC & 0.484 & 0.734 & 0.734 & 0.398 & 0.469 & 0.496 & 0.761 & 0.712 & \underline{0.764} & \textbf{0.782} \\
    MSL\_P-14   & AUPRC & 0.042 & 0.309 & 0.309 & 0.072 & 0.076 & 0.064 & 0.303 & \textbf{0.521} & 0.372 & \underline{0.449} \\
                & F1    & 0.064 & 0.502 & 0.502 & 0.182 & 0.115 & 0.104 & 0.406 & \textbf{0.639} & 0.545 & \underline{0.599} \\
          \midrule
                & AUROC & 0.500 & 0.572 & 0.580 & 0.504 & 0.527 & 0.537 & 0.641 & 0.572 & \textbf{0.701} & \underline{0.683} \\
    MSL\_P-15   & AUPRC & \textbf{0.339} & 0.017 & 0.016 & 0.014 & 0.014 & 0.015 & 0.018 & \underline{0.150} & 0.022 & 0.019 \\
                & F1    & 0.021 & 0.072 & 0.070 & 0.053 & 0.053 & 0.072 & 0.067 & \textbf{0.272} & \underline{0.087} & 0.079 \\
          \bottomrule
\end{tabular}
\end{table*}
\section{Experiments}
\subsection{Experimental Setup}
\textbf{Datasets. }
We demonstrate the effectiveness of CAROTS on five widely used real-world MTSAD datasets - SWaT~\cite{SWaT}, WADI~\cite{wadi}, PSM~\cite{PSM}, SMD~\cite{SMD}, and MSL~\cite{MSL} - and two synthetic datasets - VAR and Lorenz96~\cite{Lorenz96}.
The real-world datasets are characterized by the intricate causal relationships among variables, which are critical for understanding and detecting anomalies. For example, in SWaT and WADI, the behavior of actuators, such as pumps and valves, is directly influenced by sensor readings such as flow rates and water levels.
The two synthetic datasets are generated based on explicit linear and non-linear causal relationships. 
Following the approach of~\citet{NeurIPS-TS}, test sets for these datasets are constructed by injecting four different types of synthetic anomalies, enabling evaluation of the model performance under diverse anomaly scenarios and difficulties. Detailed descriptions of each dataset are provided in the Appendix.

\textbf{Baselines and evaluation metrics. }
We select three representative reconstruction-based models - AnomalyTransformer~\cite{AnomalyTransformer}, TimesNet~\cite{TimesNet}, and USAD~\cite{USAD} - and five contrastive learning-based models - SimCLR~\cite{SimCLR}, SSD~\cite{ssd}, CSI~\cite{csi}, CTAD~\cite{kim2023contrastive}, and CARLA~\cite{darban2025carla} - as strong baselines for comparison with CAROTS.
The MTSAD performance is evaluated using AUROC, AUPRC, and F1 scores.

\textbf{Implementation details. }
We use 20\% of the training data as validation data and apply standard normalization to the entire dataset using the mean and standard deviation of training data. We construct training, validation, and test sets using a sliding window; windows containing at least one anomaly point are labeled as anomalies. 
Unless specified otherwise, we train models with a window size of 10 and a batch size of 256 for 30 epochs.
We report the average value for each metric obtained over three random seeds. 
For contrastive learning-based models including CAROTS, we primarily employ an LSTM encoder~\cite{lstm} following CTAD~\cite{kim2023contrastive} and set the temperature parameter to 0.1. The threshold for similarity filtering is initialized to 0.5 and linearly increased to 0.9.

\subsection{Results on Real-world Datasets}
The results in Table~\ref{tab:main} showcase the effectiveness of CAROTS and CAROTS$^\dag$ (without $\mathcal{A}_{\mathrm{CD}}$) across various real-world MTSAD datasets. 
CAROTS consistently achieves the highest detection scores in nearly all metrics and datasets, while CAROTS$^\dag$ comes in a close second.
Both CAROTS and CAROTS$^\dag$ surpassing other baselines underscore that $\mathcal{A}_{\mathrm{CL}}$ from causality-aware contrastive learning is a robust MTSAD score.
For instance, CAROTS$^\dag$ already outperforms other baselines on the particularly challenging WADI and MSL\_P-15 datasets, highlighting that it is capable of identifying anomalies in complex environments.
Yet, the slight superiority of CAROTS to CAROTS$^\dag$ implies that $\mathcal{A}_{\mathrm{CD}}$ makes CAROTS even more robust.
Notably, in SWaT, CAROTS, with an AUROC of 0.852, AUPRC of 0.764, and F1 score of 0.791, outperforms all competitors.
We report full results including standard deviations in the Appendix. 

\subsection{Results on Synthetic Datasets}
In Table~\ref{tab:synthetic}, We compare CAROTS against various models on Lorenz96 and VAR, synthetically designed to exhibit clear causal relationships. 
These datasets provide a controlled environment to evaluate how effectively models leverage causal relationships for MTSAD. 
On the Lorenz96 dataset, CAROTS excels in detecting Point Global (PG) and Point Contextual (PC) anomalies, achieving AUROC scores of 0.998 and 0.975, respectively, significantly outperforming all other models. 
It also achieves the highest AUROC for Collective Global (CG) anomalies, highlighting its ability to capture both local and global causal relationships. 
On the VAR dataset, CAROTS is particularly effective at detecting Point Contextual (PC) and Collective Global (CG) anomalies with AUROC scores of 0.648 and 0.997, respectively.

\begin{table*}[t]
\centering
\footnotesize
\caption{Evaluation on synthetic datasets with explicit causal relationships: Lorenz96 and VAR. We report separate and averaged AUROC scores across four anomaly types: Point Global (PG), Point Contextual (PC), Collective Trend (CT), and Collective Global (CG).}
\label{tab:synthetic}
\vspace{0.5em}
\begin{tabular}{l||ccccc||ccccc}
\toprule
         & \multicolumn{5}{c||}{Lorenz96} & \multicolumn{5}{c}{VAR} \\
         & PG & PC & CT & CG & AVG. & PG & PC & CT & CG & AVG. \\
\midrule
AT       & 0.500 & 0.500 & 0.500 & 0.500 & 0.500 & 0.500 & 0.500 & 0.615 & 0.526 & 0.535 \\
TimesNet & 0.710 & 0.594 & \textbf{0.880} & 0.617 & 0.700 & 0.635 & 0.610 & 0.970 & \textbf{0.997} & 0.803 \\
USAD     & 0.667 & 0.601 & 0.860 & 0.506 & 0.659 & \textbf{0.663} & 0.610 & \textbf{0.968} & \textbf{0.997} & 0.810 \\
SimCLR   & 0.505 & 0.506 & 0.648 & 0.531 & 0.548 & 0.525 & 0.515 & 0.838 & 0.854 & 0.683 \\
SSD      & 0.504 & 0.507 & 0.641 & 0.526 & 0.545 & 0.525 & 0.516 & 0.837 & 0.853 & 0.683 \\
CSI      & 0.500 & 0.509 & 0.628 & 0.527 & 0.541 & 0.520 & 0.516 & 0.837 & 0.850 & 0.681 \\
CTAD     & 0.511 & 0.512 & 0.618 & 0.486 & 0.532 & 0.527 & 0.522 & 0.897 & 0.842 & 0.697 \\
\midrule
CAROTS   & \textbf{0.998} & \textbf{0.975} & 0.874 & \textbf{0.788} & 0.909 & 0.612 & \textbf{0.648} & 0.963 & \textbf{0.997} & 0.805 \\
\bottomrule
\end{tabular}
\end{table*}
\begin{figure}[t]
    \centering
    \includegraphics[width=\linewidth]{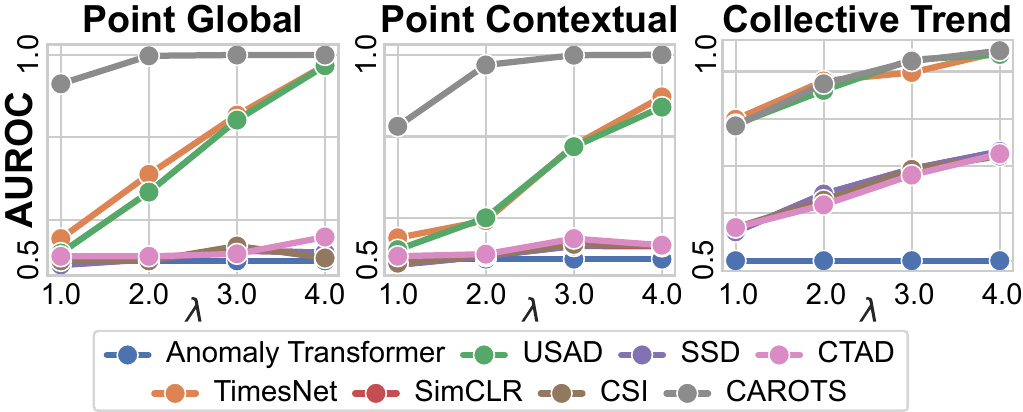}
    \vspace{-1.5em}
    \caption{Anomaly detection performance measured by AUROC across different anomaly types in the Lorenz96 dataset, evaluated based on anomaly difficulty levels controlled by the factor $\lambda$.}
    \label{fig:synthetic}
\vspace{-1em}
\end{figure}
\figurename~\ref{fig:synthetic} shows the MTSAD performance for three anomaly types in the Lorenz96 dataset across four difficulty levels. 
$\lambda$ determines the difficulty of synthetic anomalies, where a smaller $\lambda$ corresponds to injecting more challenging anomalies. 
We highlight that CAROTS demonstrates successful dtection capability even for the most difficult anomalies ($\lambda$=1.0), where other methods struggle.

\begin{table}[t]
\footnotesize
\centering
\caption{Performance evaluation on the SWaT and PSM datasets under varying temperature ($\tau$) and batch size ($B$). ROC and PRC represent AUROC and AUPRC, respectively.}
\label{tab:hparam}
\vspace{0.5em}
\begin{tabular}{ll||ccc|ccc}
\toprule
& & \multicolumn{3}{c|}{SWaT} & \multicolumn{3}{c}{PSM} \\
 & & ROC & PRC & F1 & ROC & PRC & F1   \\
\midrule
$\tau$ & 0.1         & 0.85 & 0.76 & 0.79 & 0.73 & 0.54 & 0.53 \\
 &0.2         & 0.85 & 0.76 & 0.79 & 0.76 & 0.56 & 0.57 \\
 &0.4         & 0.84 & 0.76 & 0.80 & 0.69 & 0.47 & 0.53 \\
 &0.8         & 0.84 & 0.75 & 0.78 & 0.62 & 0.43 & 0.50 \\
 &1.6         & 0.85 & 0.77 & 0.81 & 0.61 & 0.44 & 0.46 \\
\midrule
$B$ &128         & 0.85 & 0.75 & 0.77 & 0.73 & 0.54 & 0.54 \\
&256         & 0.85 & 0.76 & 0.79 & 0.76 & 0.56 & 0.57 \\
&512         & 0.86 & 0.78 & 0.80 & 0.74 & 0.55 & 0.54 \\
&1024        & 0.84 & 0.76 & 0.79 & 0.73 & 0.54 & 0.52 \\
\bottomrule
\end{tabular}
\end{table}
\begin{table}[t]
\centering
\footnotesize
\caption{Ablation studies on the SWaT dataset, evaluating the impact of key components in the CAROTS framework, including CPA, CDA, similarity filtering, and score ensemble.}
\label{tab:ablation}
\vspace{0.5em}
\begin{tabular}{l||ccc}
\toprule
             & AUROC & AUPRC & F1 \\
\midrule
w/o CPA      & 0.850     & 0.721     & 0.775  \\
w/o CDA      & 0.842     & 0.740     & 0.786  \\
w/o similarity filtering      & 0.819     & 0.733     & 0.769  \\
w/o $\mathcal{A}_{\mathrm{CL}}$ & 0.814     &  0.722    & 0.769  \\
w/o $\mathcal{A}_{\mathrm{CD}}$ & 0.861 & 0.760 & 0.789 \\
\midrule
CAROTS       & 0.852     & 0.764     & 0.792 \\
\bottomrule
\end{tabular}
\end{table}
\subsection{Analysis of CAROTS}
\textbf{Robustness to hyperparameters. }
Table~\ref{tab:hparam} analyzes the robustness of CAROTS to variations in the temperature hyperparameter ($\tau$) and batch size ($B$), both critical in contrastive learning. 
CAROTS shows stable performance across a wide range of $\tau$.
This suggests that CAROTS is resilient to changes in how similarity scores are scaled.
Similarly, CAROTS shows consistent performance across varying batch sizes, with optimal results observed in the range of $B = 256$ to $B = 512$. 
The stability across these hyperparameters implies that CAROTS can be practically and reliably applied in real-world scenarios.

\textbf{Ablation studies.} 
Table~\ref{tab:ablation} studies the effect of key technical components in CAROTS.
Without CPA, AUPRC noticeably drops from 0.764 to 0.721 and the F1 score declines from 0.792 to 0.775, which implies that observing augmented normal variations during training was crucial in CAROTS.
Similarly, the absence of SOC results in a significant decrease in AUROC from 0.852 to 0.819 and F1 score from 0.792 to 0.769. 
The contrastive learning-based anomaly score ($\mathcal{A}_{\mathrm{CL}}$) is shown to play an essential role as the primary anomaly score, as its exclusion leads to one of the largest performance drops, reducing AUROC from 0.852 to 0.814.
When CDA is removed, all variables are selected for perturbation without performing DFS, and the results indicate that CDA, along with ensemble scoring, contributes to additional robustness. 
The contrastive learning-based anomaly score ($\mathcal{A}_{\mathrm{CL}}$) is also essential, as excluding it leads to one of the largest performance drops, reducing AUROC from 0.852 to 0.814. 

\textbf{Choice of encoder architecture. }
Table~\ref{tab:arch} presents the results of CAROTS with various encoder architectures, ranging from representative recurrent architectures~\cite{lstm, gru} to a recent Transformer-based architecture~\cite{liu2024itransformer}. 
While GRU achieves the highest overall performance, the difference in performance from LSTM, GRU, and iTransformer is small.
Such a result suggests that the strong MTSAD capability of CAROTS is not contingent on a specific choice of architecture, and thus, CAROTS is compatible with different temporal modeling approaches.

\begin{table}[t]
\footnotesize
\centering
\setlength{\tabcolsep}{4pt}
\caption{Results on how CAROTS perform under different causal discovery methods. We report the averaged AUROC for runs with three different seeds.}
\label{tab:causal_discoverer}
\vspace{0.5em}
\begin{tabular}{l||cccccc}
\toprule
             & SWaT & WADI & SMD\_2-1 & SMD\_3-7 & MSL\_P-15 \\
\midrule
NGC      & 0.85    & 0.49     & 0.68 & 0.69 & 0.76   \\
CUTS   & 0.86     & 0.49     & 0.73 & 0.69 & 0.66 \\
CUTS+    & 0.85     & 0.50     & 0.70 & 0.77 & 0.70 \\
\bottomrule
\end{tabular}
\end{table}
\textbf{Choice of causal discovery model. }
Table~\ref{tab:causal_discoverer} presents results of CAROTS with different causal discovery models: NGC~\cite{tank2021neural}, CUTS~\cite{yuxiao2023cuts}, and CUTS+~\cite{cheng2024cuts+}. We find that CAROTS consistently performs well across all causal discovery methods, with only modest performance variation. While each method performs best on different datasets (e.g., CUTS+ on WADI and SMD\_3-7; CUTS on SWaT and SMD\_2-1; NGC on MSL\_P-15), the overall performance remains robust and competitive. This indicates that CAROTS does not overly depend on a particular discovery algorithm or exact causal graph structure. 
Instead of solely relying on the causal graph searched by a causal discoverer for anomaly detection, CAROTS uses the causal graph as a guide for generating semantically meaningful causality-preserving or disturbing augmentations for contrastive learning.

\begin{table}[t]
\centering
\footnotesize
\caption{Anomaly detection performance of CAROTS with different encoder architectures on the SWaT and PSM datasets.}
\label{tab:arch}
\vspace{0.5em}
\begin{tabular}{l||ccc|ccc}
\toprule
             & \multicolumn{3}{c|}{SWaT} & \multicolumn{3}{c}{PSM} \\
             & ROC   & PRC   & F1    & ROC   & PRC   & F1    \\
\midrule
LSTM         & 0.85 & 0.76 & 0.79 & 0.78 & 0.60 & 0.60 \\
GRU          & 0.86 & 0.78 & 0.81 & 0.82 & 0.63 & 0.65 \\
iTransformer & 0.85 & 0.75 & 0.78 & 0.81 & 0.60 & 0.63 \\
\bottomrule
\end{tabular}
\end{table}
\begin{figure}[t]
    \centering
    \includegraphics[width=\linewidth]{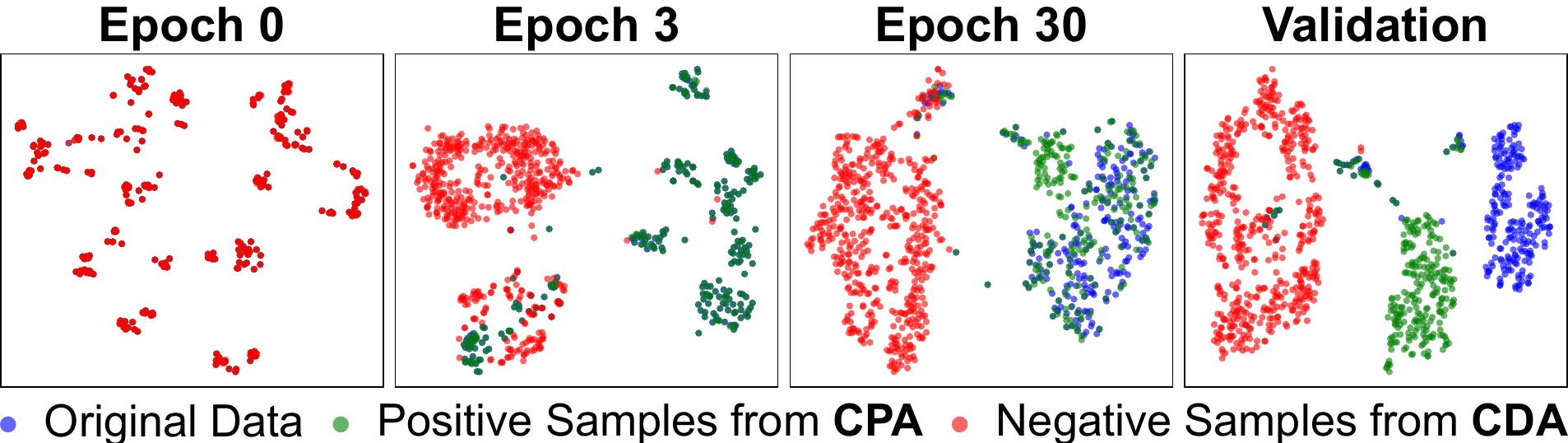}
    \vspace{-1.5em}
    \caption{T-SNE visualizations of embeddings during training and validation after training.}
    \label{fig:tsne}
\vspace{-1em}
\end{figure}
\subsection{Qualitative Analysis with t-SNE Visualization}
\figurename~\ref{fig:tsne} presents the t-SNE plots~\cite{t-SNE} of learned embeddings during CAROTS training. 
Blue, green, and red points represent original samples, positive samples from CPA, and negative samples from CDA, respectively. 
Prior to training, the data embeddings lie according to natural clusters based on the semantic similarity; thus, they show a significant overlap and appear inseparable from a causal viewpoint.
As training progresses, positive and negative samples begin to separate, showing that the proposed causality-aware contrastive learning successfully distinguishes the two.
Also, the embeddings of positive samples form distinct clusters, showing that the model effectively encodes information about different semantic patterns in the data.
In final iterations, the positive samples become more integrated, focusing on shared causal relationships.
CPA-generated samples lie outside the original data distribution, providing diverse extrapolated information that enriches the training process and enhances the model’s understanding of causal structures.
The t-SNE plot from validation data again shows a clear separation of positive and negative samples, confirming the encoder’s ability to extend causality-aware discrimination to unseen data.

\section{Conclusion}
In this study, we proposed CAROTS, a novel framework that enables causality-aware multivariate time-series anomaly detection (MTSAD).
With its unique causality-preserving and -disturbing augmentation schemes, CAROTS performs contrastive learning to train an encoder that distinguishes normal operations from anomalies based on the causal structure.
CAROTS replaces the vanilla contrastive loss with a similarity-filtered one-class contrastive loss to preserve the diversity of semantic patterns in normal operations.
The strong empirical results of CAROTS across an array of real-world and synthetic datasets validate that the consideration of causal relationships indeed equips CAROTS with enhanced anomaly detection capabilities. 
Potential directions of future research include extending CAROTS to handle more complex causal scenarios, such as confounding or exogenous variables, or non-stationary causal relationships. 

\section*{Acknowledgements}
This work was supported by Institute of Information \& communications Technology Planning \& Evaluation (IITP) grant funded by the Korea government (MSIT) [RS-2022-II220959; No.RS-2021-II211343, Artificial Intelligence Graduate School Program (Seoul National University)],
the National Research Foundation of Korea (NRF) grant funded by the Korea government (MSIT)
(No. 2022R1A3B1077720; 2022R1A5A708390811), the BK21 FOUR program of the Education
and the Research Program for Future ICT Pioneers, Seoul National University in 2025, Hyundai Motor Company, and a grant from Yang Young Foundation.

\section*{Impact Statement}
Effective anomaly detection in multivariate time series requires an understanding of the causal relationships among variables, as anomalies often disrupt these stable inter-variable dependencies rather than simply deviating from statistical norms. CAROTS pioneers a causality-aware contrastive learning framework that leverages both causality-preserving and causality-disturbing augmentations to better differentiate normal variations from true anomalies. 
This approach highlights the growing importance of causality in time-series analysis and offers a promising direction for further research—such as extending to dynamic causal structures, integrating latent confounders, or adapting to non-stationary environments—paving the way toward more robust and reliable anomaly detection systems.

\bibliography{references}
\bibliographystyle{icml2025}

\newpage
\appendix
\onecolumn
\section{Dataset Description}
\subsection{Real-world Datasets}
The datasets SWaT~\cite{SWaT}, WADI~\cite{wadi}, PSM~\cite{PSM}, MSL~\cite{MSL}, and SMD~\cite{SMD} are widely used in time series anomaly detection research, each tailored to specific domains such as industrial control systems, equipment maintenance, and environmental monitoring. These datasets provide labeled multivariate time series data and are designed to test the effectiveness of anomaly detection methods under real-world conditions. Their diverse domains and unique characteristics make them valuable resources for developing and benchmarking state-of-the-art algorithms.

The Secure Water Treatment (SWaT) dataset, developed by iTrust at the Singapore University of Technology and Design, simulates a six-stage water treatment process within an industrial control system (ICS). It consists of time series data collected from sensors and actuators, featuring both normal and attack data, where attacks simulate intentional anomalies introduced to disrupt the system. The multivariate nature of the dataset, combined with its complex sensor-actuator relationships, makes it a benchmark for ICS anomaly detection. Similarly, the Water Distribution (WADI) dataset, also created by iTrust, extends the scope to urban water distribution systems. It provides a larger and more complex simulation of long-term water network operations, capturing anomalies within a broader and more intricate system compared to SWaT.

The Proactive System Maintenance (PSM) dataset focuses on anomaly detection in industrial equipment maintenance. Derived from real-world industrial systems, this dataset contains multivariate time series data from various sensors monitoring equipment states and processes. PSM emphasizes fault detection and proactive maintenance, where anomalies represent potential equipment failures. The strong interdependencies among sensors in this dataset make it particularly useful for testing models that require context-aware anomaly detection.

In the domain of space exploration, the Mars Science Laboratory (MSL) dataset is derived from telemetry data collected by NASA's Curiosity Rover. It contains labeled normal and anomalous data based on real incidents recorded during the rover's operations. The dataset includes multivariate time series data from various sensors, reflecting the complexity and interconnectivity of space exploration systems. 

Finally, the Server Machine Dataset (SMD) represents time series data from large-scale server systems. It includes measurements from server components such as CPU, memory, and disk, along with labeled normal and anomalous states. SMD is particularly relevant for IT infrastructure and server monitoring, as it simulates realistic operational anomalies in server clusters.

\begin{table}[h]
\centering
\caption{Summary of dataset characteristics, including the number of variables, training and test sequence lengths, anomaly ratio, and stationarity assessment using Augmented Dickey-Fuller (ADF) test statistics.}
\label{tab:data}
\begin{tabular}{ccccccc}
\toprule
Dataset   & Variables & Train Steps & Test Steps & Anomaly Ratio & ADF Statistics & $p$-value \\
\midrule
SWaT      & 51  & 495,000  & 449,919  & 0.121  & -0.780  & 0.83  \\
WADI      & 123 & 784,537  & 172,801  & 0.058  & -0.347  & 0.92  \\
PSM       & 25  & 132,481  & 87,841   & 0.278  & -3.493  & 0.01  \\
SMD\_2-1  & 38  & 23,693   & 23,694   & 0.049  & -13.649 & 0.00  \\
SMD\_3-7  & 38  & 28,705   & 28,705   & 0.015  & 0.128   & 0.97  \\
MSL\_P-14 & 55  & 2,880    & 6,100    & 0.030  & -1.228  & 0.66  \\
MSL\_P-15 & 55  & 3,682    & 2,856    & 0.007  & -0.621  & 0.87  \\
\bottomrule
\end{tabular}
\end{table}

Table \ref{tab:data} summarizes the key characteristics of the datasets used in our experiments. The datasets vary in the number of variables, the length of training and test sequences, and the proportion of anomalies. The anomaly ratio represents the fraction of anomalous data points in the test set, highlighting the imbalance in certain datasets. Additionally, we report the Augmented Dickey-Fuller (ADF) test~\cite{ADFtest} statistics and corresponding p-values to assess the stationarity of each dataset, where higher p-values indicate non-stationary time-series.

For SMD and MSL, we selected subsets that exhibit stronger non-stationarity, making them more challenging for anomaly detection. These subsets better reflect real-world scenarios where complex temporal dependencies and distribution shifts pose difficulties for standard detection methods.

\subsection{Synthetic Datasets}
The Lorenz96 system is a chaotic dynamical system widely used for modeling time-series with complex temporal dependencies~\cite{Lorenz96}. Originally designed to represent simplified atmospheric dynamics, it has become a benchmark for evaluating causality-aware models, including those based on Granger causality. The system consists of $N$ coupled variables $x^i$ evolving according to the differential equation:
\begin{equation}
\frac{dx^i_t}{dt} = (x^{i+1}_t - x^{i-2}_t) x^{i-1}_t - x^i_t + F, \quad i = 1, 2, \dots, N,
\end{equation}
where $F$ is an external forcing parameter controlling the level of chaos. The indices follow cyclic boundary conditions such that $x^{-1}_t = x^{N-1}_t, \, x^0_t = x^N_t, \, x^{N+1}_t = x^1_t.$ 
From a Granger causality perspective, the Lorenz96 system exhibits strong directional dependencies between variables. Each variable $x^i$ is influenced by its two preceding variables ($x^{i-1}$ and $x^{i-2}$) and its one succeeding variable ($x^{i+1}$). We set $F=10.0$ following the values in~\cite{cheng2024cuts+}. 

The VAR dataset is generated using a Vector Autoregressive model. The VAR model captures linear dependencies among multiple time-series variables by modeling each variable as a function of both its past values and the past values of other variables in the system. A VAR model is defined as:
\begin{equation}
\bm{x}_t = \sum_{i=1}^{p} \bm{A} \bm{x}_{t-i} + \bm{\epsilon}_t,
\end{equation}
where $\bm{x}_t \in \mathbb{R}^N$ denotes the value of $N$ variables observed at time $t$, $\bm{A}$ is coefficient matrices that define the relationships between past and current values, and $p$ is the lag order, determining how many past time steps are considered. $\bm{\epsilon}_t \sim \mathcal{N}(0, \sigma^2 I)$ represents a multivariate Gaussian noise. We set $\bm{A}$, $p$, and $\sigma$ following the values in~\cite{cheng2024cuts+}.

Following the approach of~\citet{NeurIPS-TS}, we constructed the synthetic datasets with anomalies by injecting four types of synthetic anomalies, each designed to evaluate different aspects of anomaly detection. The following introduces how we generated each type of synthetic anomaly: Point Global (PG), Point Contextual (PC), Collective Trend (CT), and Collective Global (CG). 

For Point Global (PG) anomalies, anomalous time steps were randomly selected based on a predefined anomaly ratio, and affected variables were also chosen randomly. For each selected variable $x^i$, the global mean and global standard deviation were computed as:
\begin{equation}
\mu_{\text{global}} = \frac{1}{T} \sum_{t=1}^{T} x^i_t, \quad
\sigma_{\text{global}} = \sqrt{\frac{1}{T} \sum_{t=1}^{T} (x^i_t - \mu_{\text{global}})^2}.
\end{equation}

The value at the anomalous time step $t_a$ was then replaced as follows:
\begin{equation}
x^i_{t_a} = \mu_{\text{global}} + \lambda \cdot \sigma_{\text{global}},
\end{equation}

where the factor $\lambda$ controls the magnitude of deviation, allowing adjustment of anomaly difficulty.

For Point Contextual (PC) anomalies, the approach was similar, but instead of using global statistics, local mean and local standard deviation were computed over a surrounding window of radius $r$ = 5:
\begin{equation}
\mu_{\text{local}} = \frac{1}{2r+1} \sum_{t=t_a-r}^{t_a+r} x^i_t, \quad
\sigma_{\text{local}} = \sqrt{\frac{1}{2r+1} \sum_{t=t_a-r}^{t_a+r} (x^i_t - \mu_{\text{local}})^2}
\end{equation}.

The anomaly was then injected as:
\begin{equation}
x^i_{t_a} = \mu_{\text{local}} + \lambda \cdot \sigma_{\text{local}},
\end{equation}
ensuring that the anomaly is context-dependent rather than an absolute deviation.

For Collective Trend (CT) anomalies, values within the selected radius $r$ = 5 were modified by adding a linear trend with a randomly assigned slope:
\begin{equation}
x^i_t = x^i_t + \mathrm{sign} \cdot \lambda \cdot \left({t - (t_a - r)}\right), \quad t \in [t_a - r, t_a + r],
\end{equation}
where $\mathrm{sign}$ is randomly chosen as $+1$ or $-1$ ensuring that the trend can either increase or decrease, introducing deviations from the expected behavior.

For Collective Global (CG) anomalies, values within a predefined radius $r$ = 5 around the selected time step were replaced using a square sine wave, defined as:
\begin{equation}
    x^i_t = x^i_t + \sum_{k=0}^{L-1} \frac{1}{2k + 1} A \sin \left( 2\pi f (2k + 1) t \right), \quad t \in [t_a - r, t_a + r],
\end{equation}
where $A$ and $f$ control the amplitude and frequency of the anomaly. We used $A = 1.5, \, f=0.04, L=5$ following~\cite{NeurIPS-TS}. As with other anomaly types, only a subset of variables was affected.

In this experiment, synthetic datasets were generated with $N=128$ and a total length of 40,000 time steps, which were split into 16,000, 4,000, and 20,000 steps for the train, validation, and test sets, respectively. 
Following the unsupervised time-series anomaly detection setting, synthetic anomalies were injected only into the test set, with an anomaly ratio of 0.01. Among the 128 variables, 10 were randomly selected, and anomalies were introduced only into these variables. The factor $\lambda$ was set to 2.0 by default for all anomaly types except Collective Global (CG), which does not require a factor. 

\begin{table*}[t]
\footnotesize
\centering
\caption{Standard deviations for evaluation on widely-used real-world MTSAD datasets. AT and TN denote AnomalyTransformer and TimsNet, respectively. CAROTS$^\dag$ refers to the CAROTS without ensemble scoring, which only employs contrastive learning-based anomaly score.}
\label{tab:std}
\vspace{0.5em}
\begin{tabular}{ll||cccccccc||cc}
\toprule
          &       & AT    & TN & USAD  & SimCLR & SSD   & CSI   & CTAD & CARLA & CAROTS & CAROTS$^\dag$      \\
          \midrule
SWaT      & AUROC & 0.001 & 0.002    & 0.000 & 0.015  & 0.015 & 0.054 & 0.031 & 0.034 & 0.008                             & 0.003 \\
          & AUPRC & 0.013 & 0.005    & 0.000 & 0.008  & 0.008 & 0.002 & 0.160 & 0.015 & 0.003                             & 0.020  \\
          & F1    & 0.000 & 0.002    & 0.000 & 0.000  & 0.000 & 0.014 & 0.075 & 0.022 & 0.008                             & 0.009 \\
          \midrule
WADI      & AUROC & 0.025 & 0.005    & 0.003 & 0.068  & 0.068 & 0.095 & 0.016 & 0.056 & 0.007                             & 0.042 \\
          & AUPRC & 0.002 & 0.000    & 0.000 & 0.022  & 0.022 & 0.019 & 0.033 & 0.047 & 0.001                             & 0.021 \\
          & F1    & 0.000 & 0.001    & 0.002 & 0.031  & 0.031 & 0.019 & 0.043 & 0.058 & 0.002                             & 0.076 \\
          \midrule
PSM       & AUROC & 0.000 & 0.021    & 0.015 & 0.060  & 0.060 & 0.015 & 0.089 & 0.041 & 0.008                             & 0.018 \\
          & AUPRC & 0.000 & 0.010    & 0.004 & 0.050  & 0.050 & 0.012 & 0.026 & 0.012 & 0.007                             & 0.012 \\
          & F1    & 0.000 & 0.040    & 0.013 & 0.043  & 0.043 & 0.014 & 0.027 & 0.001 & 0.011                             & 0.022 \\
          \midrule
SMD\_2-1  & AUROC & 0.000 & 0.008    & 0.001 & 0.006  & 0.006 & 0.016 & 0.021 & 0.157 & 0.021                             & 0.023 \\
          & AUPRC & 0.000 & 0.007    & 0.001 & 0.017  & 0.017 & 0.023 & 0.050 & 0.078 & 0.016                             & 0.018 \\
          & F1    & 0.000 & 0.005    & 0.002 & 0.026  & 0.026 & 0.027 & 0.055 & 0.087 & 0.025                             & 0.026 \\
          \midrule
SMD\_3-7  & AUROC & 0.036 & 0.006    & 0.001 & 0.006  & 0.006 & 0.044 & 0.004 & 0.075 & 0.011                             & 0.045 \\
          & AUPRC & 0.235 & 0.015    & 0.000 & 0.050  & 0.050 & 0.135 & 0.001 & 0.050 & 0.015                             & 0.008 \\
          & F1    & 0.099 & 0.016    & 0.000 & 0.061  & 0.061 & 0.157 & 0.001 & 0.069 & 0.011                             & 0.012 \\
          \midrule
MSL\_P-14 & AUROC & 0.031 & 0.000    & 0.000 & 0.204  & 0.204 & 0.294 & 0.195 & 0.165 & 0.000                                 & 0.028 \\
          & AUPRC & 0.008 & 0.000    & 0.000 & 0.010  & 0.010 & 0.226 &0.229 & 0.154 & 0.000                                 & 0.030  \\
          & F1    & 0.005 & 0.000    & 0.000 & 0.030  & 0.030 & 0.295 & 0.274 & 0.113 & 0.000                                 & 0.051 \\
          \midrule
MSL\_P-15 & AUROC & 0.000 & 0.001    & 0.000 & 0.077  & 0.077 & 0.055 & 0.048 & 0.117 & 0.008                             & 0.012 \\
          & AUPRC & 0.289 & 0.000    & 0.000 & 0.007  & 0.007 & 0.005 & 0.004 & 0.115 & 0.001                             & 0.002 \\
          & F1    & 0.000 & 0.001    & 0.000 & 0.038  & 0.038 & 0.037 & 0.024 & 0.136 & 0.004                             & 0.013 \\
          \bottomrule
\end{tabular}
\end{table*}
\section{Additional Implementation Details}
\textbf{Baselines. }
We re-implement all baseline methods for a fair comparison. For the contrastive learning-based baselines, we follow the augmentation strategies of~\cite{kim2023contrastive}. 

\textbf{Causality-preserving augmentation. }
For causality-preserving augmentation, we randomly select a single causing variable $(M = 1)$. Gaussian noise with a standard deviation of 0.2 is applied at the last time step of $\bm{X}_{<t}$ to maintain stability.

\textbf{Causality-disturbing augmentation. }
Perturbations were introduced by adding a random bias to increase anomaly diversity and difficulty. The bias was randomly chosen from (-0.5, -0.4, -0.3, -0.2, -0.1, 0.1, 0.2, 0.3, 0.4, 0.5), creating anomalies ranging from easy to hard. To enhance augmentation diversity, perturbations were applied to 50\% of randomly selected time steps per variable. To enhance the diversity of the extracted subgraph, a cut-off probability $p$ of 0.1.

\textbf{Model optimization and evaluation. }
Each model was optimized using Adam optimizer~\cite{kingma2014adam}. We applied gradient clipping~\cite{gradientclip} with a maximum norm of 1.0 for training stability. The learning rate followed a cosine learning rate scheduling~\cite{cosineschedule} with warm up, starting at 0.0001 and increasing linearly over 5 epochs as a warm-up phase. 
A hyperparameter search was conducted over the learning rate in $\{0.001, 0.0003, 0.0001\}$ and weight decay in $\{0.001, 0.0001, 0.0\}$. 
The model with the lowest validation loss was selected for evaluation. Training was performed on a single NVIDIA A40 GPU. Following standard evaluation protocols, we reported the best F1-score. We excluded the point-adjusted F1~\cite{pointadjust}, which considers a segment as detected if any anomaly point within it is identified. This is because the point-adjusted F1 overestimates anomaly detection performance~\cite{kim2022towards}, and a robust detector should be capable of identifying arbitrary anomaly point effectively. 
For computational efficiency, the SWaT and WADI datasets were downsampled by a factor of 5. In Lorenz96 and VAR, the window sizes were set to 2 and 4 following~\cite{cheng2024cuts+}, respectively. L2 distance was used for anomaly scoring of $\mathcal{A}_{\mathrm{CL}}$, except for WADI and VAR, where cosine distance provided better performance. 

\begin{table}[t]
\centering
\caption{Anomaly detection performance on the SWaT dataset for various values of the similarity filtering threshold $\alpha$, with and without $\alpha$ scheduling. In the scheduled setting, $\alpha$ is linearly increased up to 0.9 over epochs.}
\label{tab:filter}
\begin{tabular}{cc||ccc}
\toprule
$\alpha$ scheduling & $\alpha$ & AUROC & AUPRC & F1    \\
\midrule
no                 & -1.0                    & 0.820 & 0.733 & 0.770 \\
no                 & -0.5                  & 0.820 & 0.734 & 0.766 \\
no                 & 0.0                     & 0.828 & 0.744 & 0.786 \\
no                 & 0.5                   & 0.857 & 0.761 & 0.778 \\
no                 & 0.75                  & 0.855 & 0.778 & 0.801 \\
                 \midrule
yes  & -1.0                    & 0.820 & 0.736 & 0.773 \\
yes                 & -0.5                  & 0.825 & 0.743 & 0.784 \\
yes                 & 0.0                     & 0.845 & 0.759 & 0.779 \\
yes                 & 0.5                   & 0.852 & 0.764 & 0.791 \\
yes                 & 0.75                  & 0.845 & 0.767 & 0.799 \\
                 \bottomrule
\end{tabular}
\end{table}
\section{Additional Experimental Results}
\textbf{Standard deviations. }
Table~\ref{tab:std} shows the standard deviations for experiments presented in Table~\ref{tab:main}, which is averaged for three random seeds.

\textbf{Effectiveness of similarity filtering. }
Table~\ref{tab:filter} presents anomaly detection performance on the SWaT dataset across different values of the similarity filtering threshold $\alpha$ and examines the impact of $\alpha$ scheduling. In the non-scheduled setting, $\alpha$ is fixed throughout training, while in the scheduled setting, $\alpha$ is linearly increased up to 0.9 over epochs. The results indicate that increasing $\alpha$ generally improves AUROC and AUPRC, suggesting that a more relaxed similarity filtering threshold enhances anomaly detection. 

Applying $\alpha$ scheduling consistently leads to better performance, particularly in AUPRC and F1 score, highlighting its role in improving model robustness. The performance gains are more pronounced at intermediate $\alpha$ values, where scheduling stabilizes results and enhances the model’s ability to distinguish anomalies. 
Overall, a moderate similarity filtering threshold combined with progressive $\alpha$ scheduling yields the best results. This underscores the importance of adaptively tuning $\alpha$ over training to balance anomaly separation and representation diversity. 

\begin{table}[t]
\footnotesize
\centering
\caption{Ablation studies on 7 real-world datasets and 2 synthetic datasets. We report the averaged AUROC for runs with three different seeds. For Lorenz96 and VAR, the reported values represent the average performance across the four different synthetic anomaly types.}
\label{tab:ablation_extended}
\vspace{0.5em}
\begin{tabular}{l||ccccccccc}
\toprule
             & SWaT & WADI & PSM & SMD\_2-1 & SMD\_3-7 & MSL\_P-14 & MSL\_P-15 & Lorenz96 & VAR \\
\midrule
w/o CPA      & 0.850     & 0.486     & 0.786 & 0.700 & 0.756 & 0.764 & 0.740 & 0.918 & 0.767  \\
w/o CDA      & 0.842     & 0.488     & 0.789 & 0.623 & 0.732 & 0.764 & 0.609 & 0.917 & 0.785  \\
w/o similarity filtering      & 0.819     & 0.493     & 0.706 & 0.758 & 0.719 & 0.764 & 0.719 & 0.923 & 0.765 \\
w/o $\mathcal{A}_{\mathrm{CL}}$ & 0.814     &  0.494    & 0.778 & 0.602 & 0.701 & 0.768 & 0.694 & 0.943 & 0.732  \\
CAROTS$^\dag$ (w/o $\mathcal{A}_{\mathrm{CD}}$) & 0.861 & 0.622 & 0.729 & 0.726 & 0.779 & 0.782 & 0.683 & 0.919 & 0.769 \\
CAROTS       & 0.852     & 0.502     & 0.783 & 0.703 & 0.769 & 0.764 & 0.701 & 0.909 & 0.805 \\
\bottomrule
\end{tabular}
\end{table}
\begin{table}[t]
\centering
\caption{Performance metrics across different $\sigma$ in CPA on SWaT dataset.}
\label{tab:ablation_sigma}
\begin{tabular}{c|c|c|c}
\toprule
$\sigma$ & AUROC & AUPRC & F1 \\
\midrule
0    & 0.850$\pm$0.001 & 0.761$\pm$0.002 & 0.798$\pm$0.001 \\
0.05 & 0.853$\pm$0.003 & 0.762$\pm$0.002 & 0.797$\pm$0.004 \\
0.1  & 0.852$\pm$0.008 & 0.764$\pm$0.003 & 0.791$\pm$0.008 \\
0.2  & 0.849$\pm$0.007 & 0.759$\pm$0.009 & 0.795$\pm$0.000 \\
0.4  & 0.848$\pm$0.002 & 0.762$\pm$0.007 & 0.792$\pm$0.001 \\
\bottomrule
\end{tabular}
\end{table}
\textbf{Extended ablation studies. }
Table~\ref{tab:ablation_extended} shows the extended ablation studies of Table~\ref{tab:ablation} on 7 real-world datasets and 2 synthetic datasets, validating the effectiveness of each component of CAROTS. Table~\ref{tab:ablation_sigma} presents the performance across different values of $\sigma$ in CPA. Performance remains stable across different $\sigma$'s, indicating that the generated samples do not degrade model quality, even with higher noise levels. These results suggest that CPA is robust to the choice of $\sigma$ within a reasonable range.

\begin{table}[t]
\centering
\caption{Comparison of cosine similarity of causality matrix across different quarters for SWaT, WADI, and PSM datasets.}
\label{tab:over_time}
\begin{tabular}{c|c|c|c}
\toprule
Quarters & SWaT & WADI & PSM \\
\midrule
Q1vsQ2 & 0.911 & 0.965 & 0.955 \\
Q1vsQ3 & 0.923 & 0.966 & 0.953 \\
Q1vsQ4 & 0.928 & 0.959 & 0.918 \\
Q2vsQ3 & 0.978 & 0.973 & 0.952 \\
Q2vsQ4 & 0.978 & 0.964 & 0.898 \\
Q3vsQ4 & 0.981 & 0.963 & 0.915 \\
\bottomrule
\end{tabular}
\end{table}
\textbf{Causal relationships over time. } 
CAROTS assumes that causal relationships among variables remain consistent over time, following existing literature on causal discovery~\cite{tank2021neural, yuxiao2023cuts, cheng2024cuts+}. To empirically assess whether this statement holds in our setting, we further analyze the evolution of causal structures in three benchmark datasets: SWaT, WADI, and PSM. For each dataset, we split the normal training data into four disjoint, time-ordered segments (quarters 1 to 4), train a causal discovery model on each, and compute pairwise cosine similarities between the resulting graphs. As shown in Table~\ref{tab:over_time}, the consistently high similarity indicates that the learned causal relationships remain stable across time segments, supporting the validity of our approach.

\begin{table}[t]
\centering
\caption{Comparison of training time for different anomaly detectors on SWaT dataset.}
\label{tab:computation}
\begin{tabular}{l|c}
\toprule
Method & Time (min) \\
\midrule
CAROTS & 25 \\
AnomalyTransformer & 12 \\
TimesNet & 56 \\
USAD / SimCLR / SSD & 6 \\
CSI / CTAD & 10 \\
\bottomrule
\end{tabular}
\end{table}
\textbf{Computational cost. }
We compare the training times of the evaluated anomaly detectors in Table~\ref{tab:computation}. The pre-training time of the causal discovery model is excluded to ensure a fair comparison of the anomaly detector training times. Even with causal modules, CAROTS is efficient due to a lightweight one-layer LSTM. The train time of CAROTS is comparable to baselines and cheaper than heavier models like TimesNet, indicating that leveraging causal structure can reduce reliance on deeper architectures.

\begin{figure}[t]
    \centering
    \includegraphics[width=\linewidth]{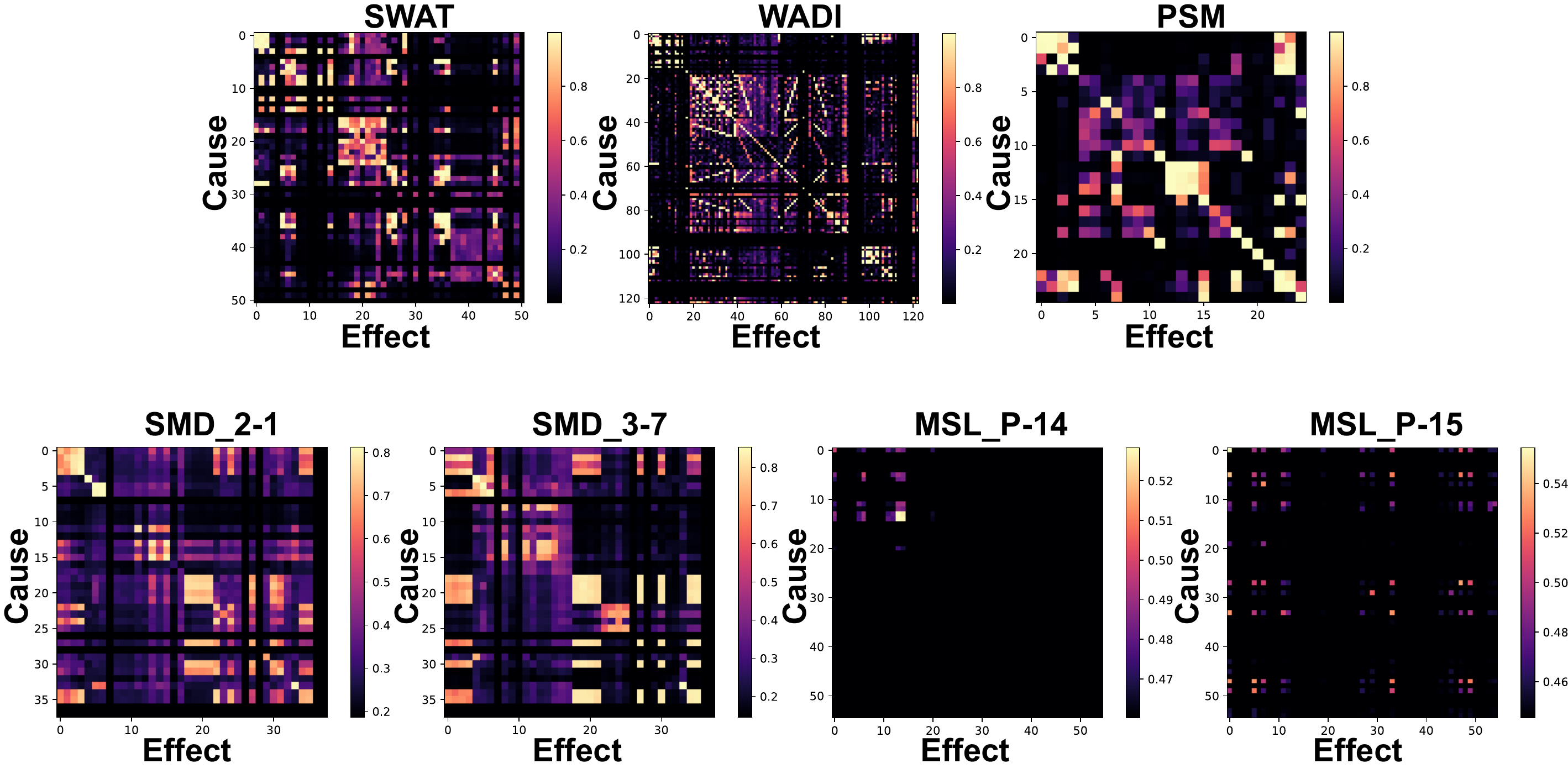}
    \vspace{-1.5em}
    \caption{Learned causality matrix from the causal discovery model for each dataset.}
    \label{fig:causality_mtx}
\end{figure}
\textbf{Learned causality matrix. }
\figurename~\ref{fig:causality_mtx} presents the learned causality matrix from the causal discovery model for each dataset.

\section{Limitations and Future Work}
CAROTS demonstrates strong performance in MTSAD by effectively leveraging causal relationships. At the same time, there remain opportunities to further enhance its robustness and applicability in more complex scenarios. First, CAROTS assumes that causal relationships in normal operations remain largely time-invariant. While this assumption holds in many stable environments, real-world systems often experience gradual or abrupt causal shifts due to evolving conditions or external interventions. Adapting CAROTS to dynamically capture and adjust to non-stationary causal structures would further strengthen its applicability to real-world settings. 
Second, CAROTS focuses on direct causal relationships but does not explicitly model confounding variables, which can introduce hidden biases. Addressing this challenge by developing mechanisms to identify and mitigate confounding effects could improve the reliability of causality-aware anomaly detection. 
Third, exogenous variables, which influence the system externally without being directly modeled, may impact the learned causal relationships. Incorporating external contextual factors, such as domain knowledge or auxiliary data sources, could expand CAROTS’ effectiveness in handling more complex causal interactions. Despite these challenges, CAROTS introduces a novel and pioneering approach to MTSAD by integrating causal awareness into contrastive learning. To the best of our knowledge, this is the first framework to explicitly incorporate causality into contrastive MTSAD, providing a new perspective on how anomalies can be characterized through deviations from stable causal structures.


\end{document}